\documentclass[letterpaper, 10 pt, conference]{ieeeconf}  

\IEEEoverridecommandlockouts                              

\usepackage{cite}
\usepackage{amssymb}
\usepackage{graphicx}
\usepackage{textcomp}
\usepackage[utf8]{inputenc}
\usepackage[T1]{fontenc}
\usepackage{url}
\usepackage{booktabs}
\usepackage{amsfonts}
\usepackage{nicefrac}
\usepackage{microtype}
\usepackage{booktabs}
\usepackage{algorithm}
\usepackage{algorithmic}
\usepackage{mathtools}
\usepackage{amsmath}
\usepackage{upgreek}
\usepackage{multicol}
\usepackage{bookmark}
\usepackage{multirow}
\usepackage{tabularx}

\usepackage{subcaption}
\usepackage[table,dvipsnames]{xcolor}
\usepackage{breqn}
\usepackage{wrapfig}
\usepackage{hyperref}

\definecolor{lightblue}{rgb}{0.8,0.9,1}
\newcolumntype{C}{>{\centering\arraybackslash}X}
\def\BibTeX{{\rm B\kern-.05em{\sc i\kern-.025em b}\kern-.08em
    T\kern-.1667em\lower.7ex\hbox{E}\kern-.125emX}}

\title{\LARGE \bf
OffLight: An Offline Multi-Agent Reinforcement Learning Framework for Traffic Signal Control
}

\author{Rohit Bokade and Xiaoning Jin
\thanks{Both authors are with the Department of Mechanical and Industrial Engineering, Northeastern University, Boston, MA 02115, USA. 
{\tt\small \{bokade.r, xi.jin\}@northeastern.edu}}%
}

\begin{document}

\maketitle
\thispagestyle{empty}
\pagestyle{empty}

\begin{abstract}

Efficient traffic signal control (TSC) is crucial for reducing congestion and improving urban mobility. While multi-agent reinforcement learning (MARL) has shown promise in adaptive traffic management, its reliance on real-time interactions makes online training costly and impractical. Offline MARL, which learns from historical data, offers a safer and more scalable alternative but struggles with heterogeneous behavior policies in real-world datasets. We introduce OffLight, a novel offline MARL framework designed to handle policy heterogeneity in traffic signal control datasets. OffLight integrates importance sampling (IS) to correct for distributional shifts and return-based prioritized sampling (RBPS) to emphasize high-quality experiences. To model diverse behavior policies, it leverages a Gaussian mixture model variational graph autoencoder (GMM-VGAE), capturing spatial and temporal traffic dynamics while improving policy estimation. We evaluate OffLight across real-world urban traffic scenarios varying from small to large traffic signal networks. Results show that OffLight outperforms existing offline RL methods, reducing average travel time by up to 7.8\% and queue length by 11.2\% compared to state-of-the-art approaches. Unlike prior offline RL methods, OffLight effectively adapts to complex and mixed-policy datasets, making it more reliable for real-world deployment without risky online training. These results highlight OffLight’s potential to improve urban traffic management at scale. Our implementation is available at \url{https://github.com/rbokade/offlight}.

\end{abstract}

\section{INTRODUCTION}

Efficient traffic signal control (TSC) is critical for urban mobility, directly affecting congestion, travel times, and city livability. Multi-agent reinforcement learning (MARL) has emerged as a promising solution by enabling decentralized, adaptive, and intelligent control of traffic signals~\cite{haydari2020deep,noaeen2022reinforcement}. While online MARL methods rely on real-time interaction, offline MARL uses pre-collected data, offering safer, more cost-efficient, and scalable alternatives by eliminating the need for risky real-time experimentation~\cite{levine2020offline,zhan2022offline,fu2020d4rl,lange2012batch}. 

\begin{figure*}[!ht]
    \centering
    \includegraphics[width=0.85\textwidth]{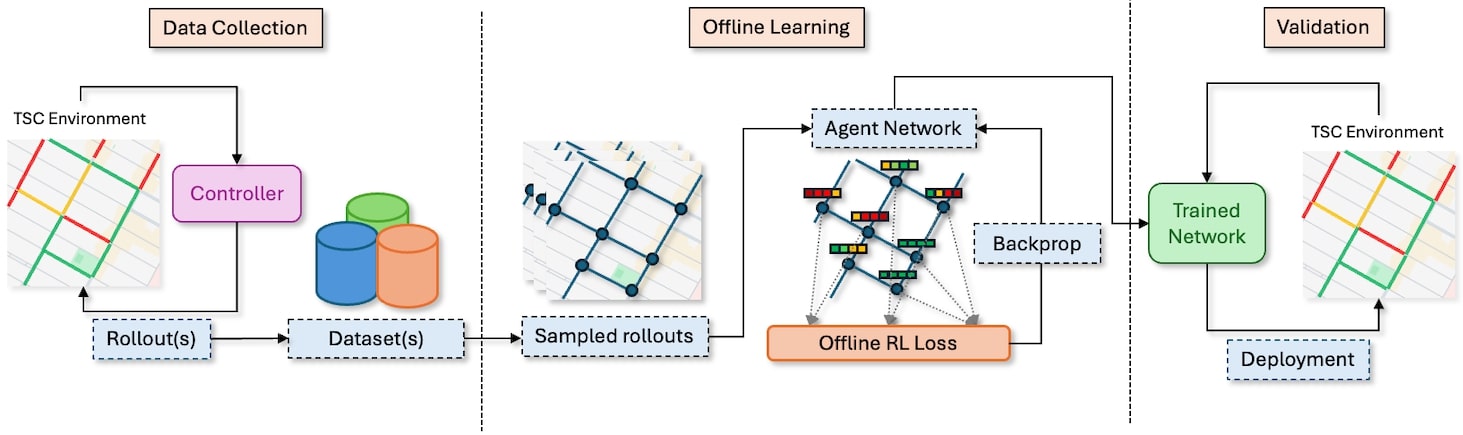}
    \caption{General Offline MARL Framework for Traffic Signal Control}
    \label{fig:offline_rl}
    \vspace{-1.5em}
\end{figure*}

Several offline RL methods have shown promise in traffic signal control by using imitation learning to bootstrap initial training. Methods such as DemoLight~\cite{xiong2019learning}, Cooperative Control~\cite{huo2020cooperative}, and CrossLight~\cite{sun2024crosslight} have leveraged expert demonstrations to quickly initialize and refine traffic control policies. More recently, DataLight~\cite{zhan2022offline} improved policy performance through attention mechanisms that capture spatial traffic distributions. However, these approaches do not fully address a significant issue in offline MARL datasets: heterogeneous behavior policies arising from varied control strategies, temporal variability, and the mixture of different operational contexts~\cite{chen2022real,wei2019colight}.

The heterogeneity in real-world traffic datasets poses significant challenges for offline MARL, primarily due to distributional shifts and suboptimal behavior biases that degrade policy learning~\cite{kumar2020conservative, wu2019behavior}. Recent studies, including Hong et al.\cite{hong2023beyond} and Yue et al.\cite{yue2023decoupled}, have proposed prioritized resampling techniques to emphasize high-quality experiences, thus improving policy robustness in offline settings. Nevertheless, effectively modeling diverse behavior policies in complex multi-agent environments remains a substantial challenge for offline MARL methods.

To address these challenges, we propose OffLight, an offline MARL framework specifically designed for handling heterogeneous behavior policies in TSC datasets. OffLight integrates importance sampling (IS) to correct distributional shifts and return-based prioritized sampling (RBPS) to highlight high-quality experiences. Additionally, OffLight uses a Gaussian mixture model variational graph autoencoder (GMM-VGAE) to represent diverse behavior policies accurately. Experiments across real-world urban traffic scenarios show that OffLight outperforms existing offline RL methods, demonstrating reduced average travel times and queue lengths. Our results highlight OffLight’s capability to reliably manage urban traffic networks without the risks associated with online training, making it a practical and scalable solution for real-world deployment. 

\section{Background}
\label{sec:background}

\subsection{Multi-Agent Reinforcement Learning}

We model the MARL problem as Decentralized Partially Observable Markov Decision Processes (Dec-POMDPs)~\cite{oliehoek2016concise}, formally defined by the tuple $\langle N, \mathcal{S}, \{\mathcal{A}^i\}_{i=1}^N, \{\mathcal{O}^i\}_{i=1}^N, P, R, H, \gamma \rangle$. Here, $N$ is the number of agents, $\mathcal{S}$ is the state space, $\mathcal{A}^i$ and $\mathcal{O}^i$ are the action and observation spaces for agent $i$, respectively, $P$ is the state transition probability, $R$ is the shared reward, $H$ is the finite horizon, and $\gamma \in [0,1)$ is the discount factor.

Each agent $i$ selects actions based on its local action-observation history $h^i_t = \{o^i_0, a^i_0, \dots, o^i_t\}$, resulting in joint action $\mathbf{a}_t = (a_t^1, \dots, a_t^N)$, transitioning the environment to state $s_{t+1}$ according to $P$, and yielding a shared reward $R$. The objective is to determine decentralized policies $\{\pi^i\}_{i=1}^N$, with each policy $\pi^i: H^i_t \rightarrow \mathcal{A}^i$ mapping histories to actions, to maximize expected cumulative reward\footnote{In practice, we employ local rewards for each agent and adopt Centralized Training with Decentralized Execution (CTDE), as we found this approach converges faster than using a scalar global reward.}:
\begin{equation}
    \max_{\{\pi^i\}} \mathbb{E}\left[\sum_{t=0}^{H}\gamma^t R(s_t,\mathbf{a}_t)\right].
\end{equation}

To optimize decentralized policies, Deep Q-Learning (DQNs) \cite{mnih2015human} is commonly employed, where each agent independently updates its Q-values based on local observations and rewards as follows:
\begin{dmath}
    Q^i(o^i_t,a^i_t) \leftarrow Q^i(o^i_t,a^i_t) + \alpha \left(r^i_t + \gamma \max_{a'} Q^i(o^i_{t+1},a') - Q^i(o^i_t,a^i_t)\right).
\end{dmath}

However, independent learning may lead to suboptimal coordination. Recent frameworks, such as CoLight \cite{wei2019colight}, PressLight \cite{wei2019presslight}, and AttendLight \cite{oroojlooy2020attendlight}, address this by incorporating Graph Attention Networks (GATs)~\cite{velivckovic2017graph} for inter-agent communication. We leverage these successful frameworks as the basis for our approach.

\subsection{Offline Multi-Agent Reinforcement Learning}

Offline Multi-Agent Reinforcement Learning (Offline MARL) learns policies from fixed datasets without additional environment interaction during training \cite{levine2020offline,zhan2022offline}. 

Offline MARL presents several challenges: \textbf{(1) Distributional Shift}: The dataset may differ significantly from the policy's actual environment interactions, leading to inaccurate value estimations and reduced performance \cite{kumar2019stabilizing}. \textbf{(2) Extrapolation Error}: Without interactive feedback, agents risk generalizing poorly to situations not covered by the dataset. \textbf{(3) Limited Coverage}: Historical data might not include optimal coordination strategies necessary for effective multi-agent policies.

Recent offline RL algorithms address these issues. Conservative Q-Learning (CQL) \cite{kumar2020conservative} introduces a penalty for overestimating Q-values of actions not in the dataset. Its objective is:
\begin{dmath}
\min_{Q_{i}} \alpha \left( \mathbb{E}{o_{i} \sim \mathcal{D}, a_{i} \sim \pi(a_{i}|o_{i})} [Q(o_{i},a_{i})] - \mathbb{E}{(o_{i},a_{i}) \sim \mathcal{D}} [Q(o_{i}, a_{i})] \right) + \mathcal{L}_{\text{Bellman}}(Q_{i}).
\end{dmath}

TD3+BC \cite{fujimoto2021minimalist} combines TD3 with a behavior cloning term to reduce policy deviation from the dataset, thus managing distributional shift:
\begin{dmath}
\min_{\pi_{\theta_{i}}} -\mathbb{E}{o_{i} \sim \mathcal{D}} [Q\phi(o_{i}, \pi_{\theta_{i}}(o_{i}))] + \lambda \mathbb{E}{(o_{i},a_{i}) \sim \mathcal{D}} [|\pi_{\theta_{i}}(o_{i}) - a_{i}|_2^2].
\end{dmath}

Both methods can be adapted to multi-agent systems using Centralized Training with Decentralized Execution (CTDE) \cite{marl-book,foerster2018counterfactual}, enabling centralized training and decentralized agent execution for better scalability and coordination.

\subsection{Motivation for the OffLight Framework}

Offline MARL faces unique challenges when datasets contain heterogeneous behavior policies, especially in domains like Traffic Signal Control (TSC). Different operational contexts or sources create diverse behaviors, increasing the likelihood of distributional mismatch. This diversity leads to unreliable Q-value estimates and degraded performance, as distributional correction methods become less effective \cite{kumar2019stabilizing}. Additionally, accurately estimating mixed behavior policies is difficult, complicating policy constraints and importance sampling \cite{wu2019behavior}.

To address heterogeneous behavior policies in offline MARL, we propose the \textbf{OffLight} framework. OffLight utilizes a behavior policy estimation technique (GMM-VGAE) to accurately model policy diversity, employs importance sampling to mitigate distributional shift, and integrates Reward-Based Prioritized Sampling (RBPS) for improved sample efficiency. RBPS prioritizes high-reward episodes, enhancing the focus on valuable training data. These strategies aim to improve policy robustness and performance specifically in Traffic Signal Control (TSC) scenarios. The OffLight framework details are introduced in the next section.

\section{OffLight Framework}
\label{sec:offlight}

\begin{figure}[!h]
    \centering
    \includegraphics[width=0.85\linewidth]{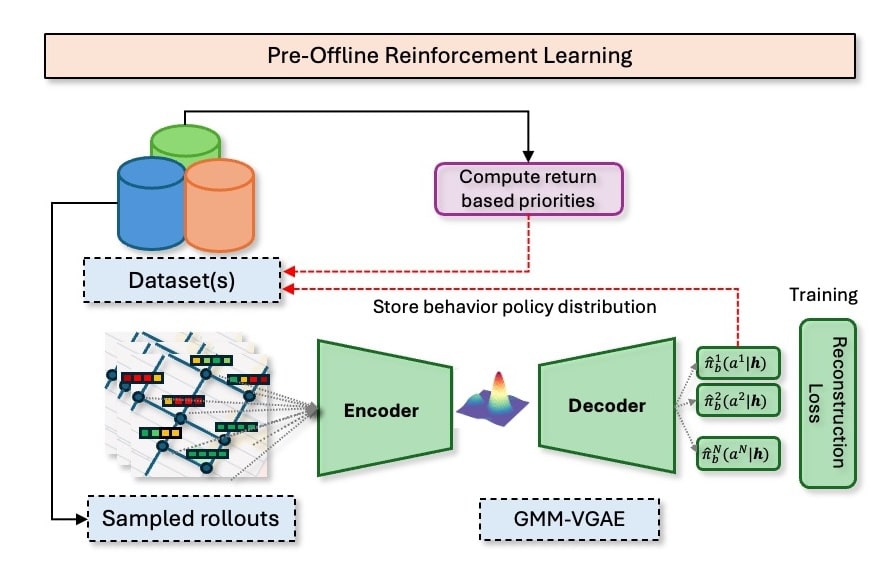}
    \caption{OffLight Architecture: Integrating GMM-VGAE with IS and RBPS for offline MARL in traffic signal control}
    \label{fig:graph_gmvae}
    \vspace{-1.25em}
\end{figure}

OffLight is designed to tackle the challenges of offline MARL in the domain of traffic signal control (TSC). It combines a self-supervised learning approach with IS to account for heterogeneous policies and improve learning from static, pre-collected datasets. IS and RBPS weights are collected before the offline RL training and stored (see Figure \ref{fig:graph_gmvae}). During the offline RL training phase, these weights are used to improve learning. Below, we outline the architecture and key components of OffLight, highlighting how it handles mixed-policy data to enhance traffic management performance.

\subsection{OffLight Agent Design}

\noindent \textbf{Observation Space:} Each traffic signal agent has a limited range of observation. This range reflects the realistic coverage of common sensors such as inductive loop detectors or cameras. For each incoming lane, the agent observes: (1) number of vehicles $\{n_{l}\}_{l=1}^{L_{i}}$, where $L_{i}$ is the number of lanes for traffic signal $i$, average speed of vehicles $\{s_{l}\}_{l=1}^{L_{i}}$, normalized by the speed limit, number of halted vehicles, representing the queue length $\{q_{l}\}_{l=1}^{L_{i}}$, and the current phase ID, indicating the traffic signal's current phase. These local observations provide the sensory input required for each agent to make real-time decisions about the traffic signal phases.

\noindent \textbf{Action Space:} Each traffic signal agent is responsible for managing the signal phases at its respective intersection. The agent's available actions involve selecting one of the predefined green phases applicable to that intersection. Once a green phase is chosen, the system enforces a mandatory transition to a yellow phase. The action selection occurs at fixed intervals of 5 simulation seconds.

\noindent \textbf{Reward Function:} We use minimizing the queue length as objective. The reward for each agent $i$ at each time step $r^{i}_{t}$ is defined as $r^{i}_{t} = -\sum_{l \in L_{i}} q_{l}(t),$ where $q_{l}(t)$ represents the queue length on lane $l$ at time $t$, and $L_{i}$ is the set of incoming lanes for traffic signal $i$. This reward structure encourages agents to minimize congestion and improve traffic flow throughout the system.

\subsection{Dataset Structure}

We consider a dataset $\mathcal{D}$ collected from a network of $N$ intersections over $M$ episodic trajectories. Each trajectory $\tau_k$ comprises a sequence of interactions between the agents (traffic signals) and the environment. Formally, the dataset is represented as:
\begin{equation}
    \mathcal{D} = \left\{ \tau_{k} \right\}_{k=1}^M, \quad \text{where} \quad \tau_{i} = \left\{ \left( \mathcal{O}_{t}, \mathcal{A}_{t}, \mathcal{O}'_{t}, \mathcal{R}_{t} \right) \right\}_{t=0}^{T_k},
\end{equation}
where $\mathcal{O}_{t} = \left\{ o_t^{1}, o_t^{2}, \dots, o_{t}^{N} \right\}$ denotes the observations at time step $t$ for each of the $N$ intersections, $\mathcal{A}_{t} = \left\{ a_{t}^1, a_{t}^2, \dots, a_{t}^{N} \right\}$ represents the joint actions taken by the agents at time $t$, $\mathcal{O}'_{t} = \left\{ o_{t+1}^1, o_{t+1}^2, \dots, o_{t+1}^{N} \right\}$ are the subsequent observations after taking actions $\mathcal{A}_t$, $\mathcal{R}_{t}$ is the reward received at time $t$, and $T_k$ is the length of trajectory $\tau_k$. This structure reflects real-world data collected from sensor-equipped intersections and enables learning traffic control policies from historical data.

\subsection{Self-Supervised Learning of Behavior Policies}

OffLight employs GMM-VGAE to model heterogeneous policies, using: \textbf{(1) GMM-VGAE for Policy Diversity:} A Gaussian mixture graph variational autoencoder model (GMM) captures distinct traffic control policies. \textbf{(2) GATs and LSTMs for Spatial-Temporal Dynamics:} Graph attention networks (GATs) encode local interactions, while LSTMs track temporal trends. 

These components construct a structured latent space for estimating policies at each intersection. The encoder processes agent observations and past actions to capture spatial-temporal dependencies. The latent space represents diverse traffic control policies using a GMM-VGAE model. Finally, the decoder reconstructs policy distributions based on the latent features and current observations. We also enable parameter sharing among agent, reducing model complexity while improving efficiency. This design allows OffLight to effectively model heterogeneous policies across intersections. 

\subsection{Importance sampling Integration}

To address the challenge of distributional shift between the behavior policy and the target policy, OffLight integrates IS. This mechanism adjusts the influence of each transition based on its alignment with the target policy, ensuring that the learning algorithm emphasizes relevant and high-quality data.
\begin{equation}
    w_{\text{IS}, t}^{k} = \frac{1}{N} \sum_{n=1}^{N} \frac{\pi_\theta^{i}(a_t^{i} \mid \mathbf{h}_{t}^{i})}{\hat{\pi}_b^{i}(a_t^{i} \mid \mathbf{h}_{t}^{i}, \mathbf{z}_{t})}
\end{equation}
where \( w_{\text{IS}, t}^{k} \) is the importance sampling weight for the transition at time step \( t \) in episode \( k \), \( \pi_\theta^{i}(a_{t}^{i} \mid \mathbf{h}_{t}^{i}) \) \footnote{\(\mathbf{h}_{t}^{i}\) is computed for each agent using Graph Neural Networks (GNNs), which aggregate information from neighboring agents.} denotes the probability of agent \( n \) taking action \( a_t^{i} \) under the learned \textbf{target policy} \( \pi_\theta \), conditioned on the neighborhood aggregated information \( \mathbf{h}_{t}^{i} \), and \( \hat{\pi}_{b}^{i}(a_t^{i} \mid \mathbf{h}_{t}^{i}, \mathbf{z}_{t}) \) represents the estimated \textbf{behavior policy probability} for agent \( n \), obtained from the GMM-VGAE, conditioned on it's hidden state \( \mathbf{h}_{t}^{i} \) and the latent embedding of behavior policy $\mathbf{z}_{t}$.

IS corrects for the discrepancy between the behavior policy and the target policy by re-weighting transitions \footnote{The \textbf{target policy} is the policy that the offline RL algorithm seeks to optimize, while the \textbf{behavior policy} refers to the policy used to generate the offline dataset.}. Transitions where the actions are more probable under the target policy relative to the behavior policy receive higher importance sampling weights.

\subsection{Return-Based Prioritized Sampling}
\label{sec:prioritized_sampling}

To enhance sample efficiency and accelerate the learning process, OffLight employs RBPS, similar to \cite{yue2023decoupled}. This strategy prioritizes episodes based on their cumulative rewards, ensuring that the learning algorithm focuses on more successful traffic control experiences.
\begin{equation}
    w_{\text{RBPS}}^{k} = C \left( \frac{G_{k} - G_{\text{min}}}{G_{\text{max}} - G_{\text{min}}} + p_{\text{base}} \right)
\end{equation}
where \( w_{\text{RBPS}}^{k} \) is the Return-Based Prioritized Sampling weight for episode \( k \), \( G_{k} = \sum_{t=0}^{T_{k}} r_{t} \) is the total reward for episode \( k \), \( G_{\text{min}} \) and \( G_{\text{max}} \) are the minimum and maximum total rewards across all episodes in the dataset, respectively. \( p_{\text{base}} \) is a small positive constant added to ensure that all episodes have a non-zero probability of being sampled, and \( C \) is a normalization constant to maintain numerical stability or ensure the weights sum to a desired value.

\begin{figure}[!h]
    \centering
    \hspace{-1.5em}
    \subfloat[Jinan]{
        \label{fig:jinan_returns_dist}
        \includegraphics[width=0.33\linewidth]{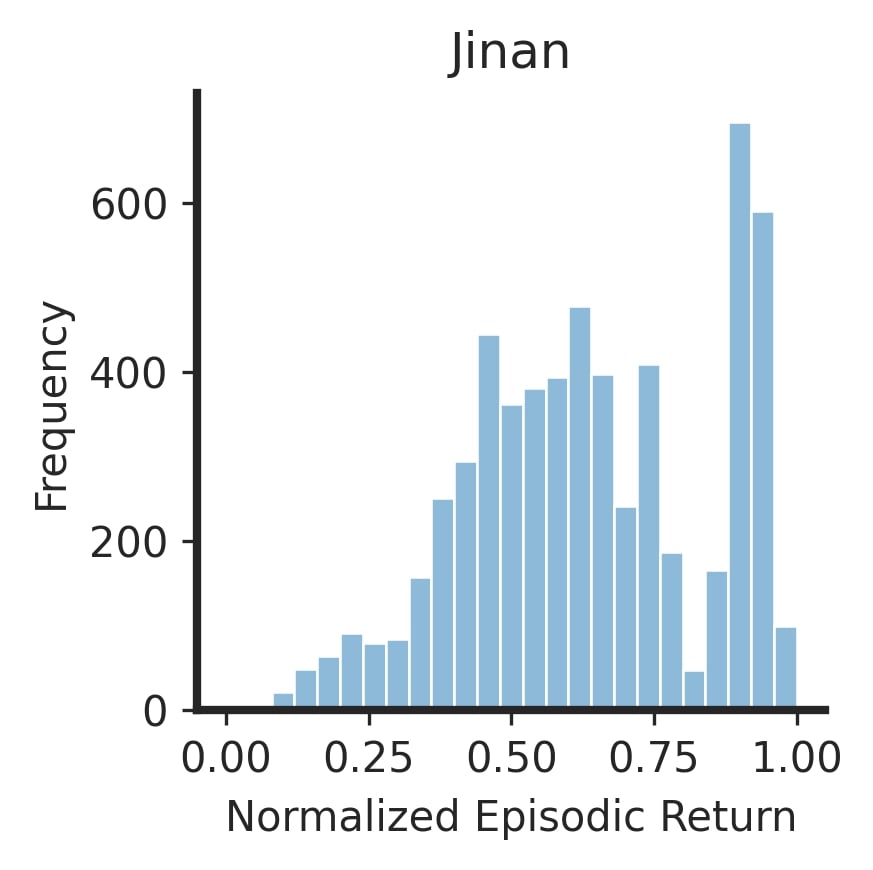}
    }
    \hspace{-1.2em}
    \subfloat[Hangzhou]{
        \label{fig:hangzhou_returns_dist}
        \includegraphics[width=0.33\linewidth]{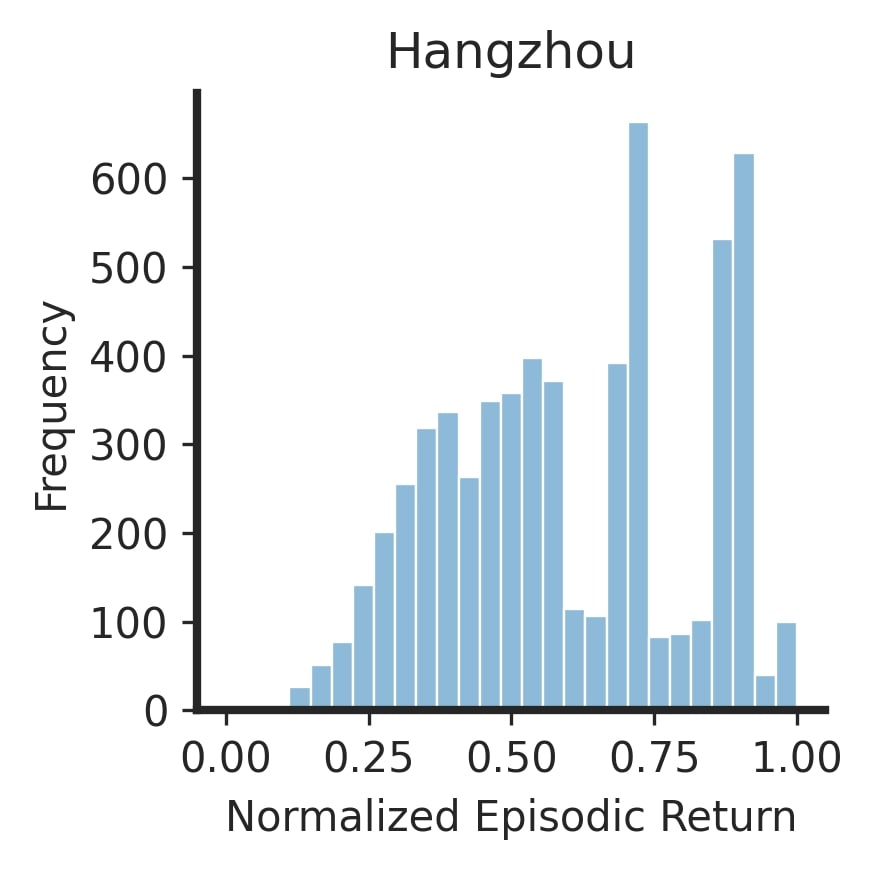}
    }    
    \hspace{-1.2em}
    \subfloat[Manhattan]{
        \label{fig:manhattan_returns_dist}
        \includegraphics[width=0.33\linewidth]{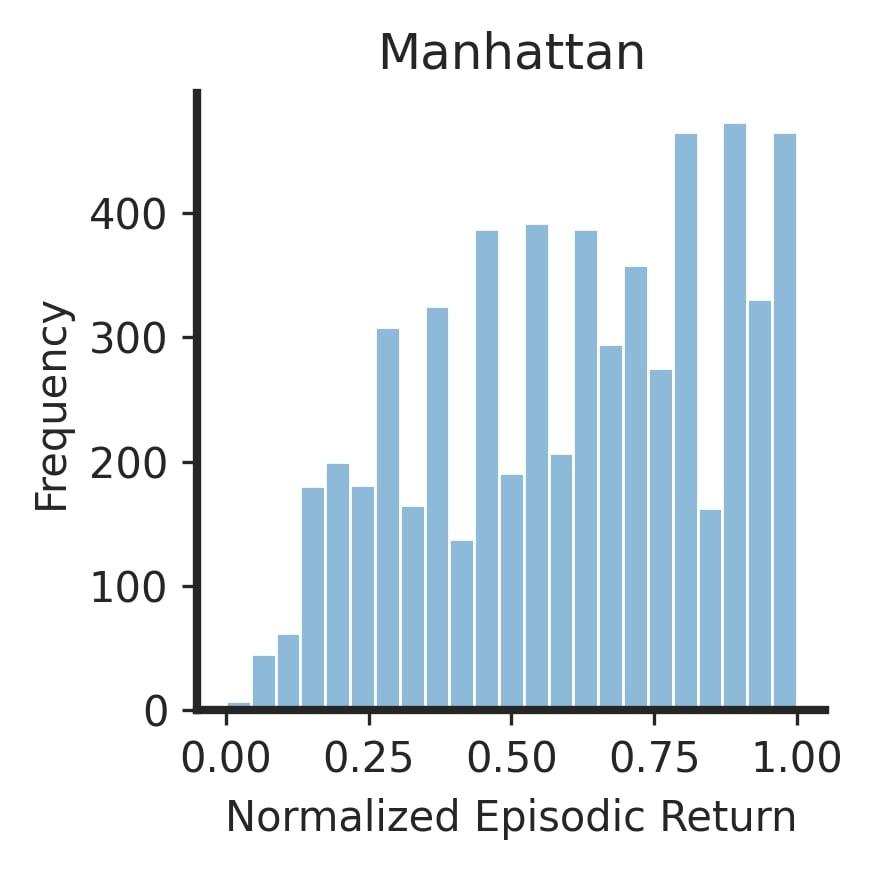}
    }    
    \caption{Distribution of Episodic Returns in mixed-policy datasets. It highlights the variability and heterogeneity in the offline dataset, with some episodes achieving high returns while others are suboptimal. RBPS targets this imbalance by prioritizing episodes with higher returns, ensuring that learning is focused on successful traffic control strategies.}
    \label{fig:return_distribution}
    \vspace{-1.5em}
\end{figure}

RBPS ensures that episodes with higher returns, indicative of more effective traffic management, are sampled more frequently. This focus accelerates learning by emphasizing experiences that contribute significantly to improved traffic flow and reduced congestion. This especially proves beneficial when the distribution of the data is skewed or multimodal as can be seen in Figure \ref{fig:return_distribution}.

\subsection{Combining importance sampling and Return-Based Prioritized Sampling}
\label{sec:combining_methods}


\begin{figure*}[!ht]
    \centering
    \subfloat[Jinan]{
        \label{fig:jinan}
        \includegraphics[width=0.25\textwidth]{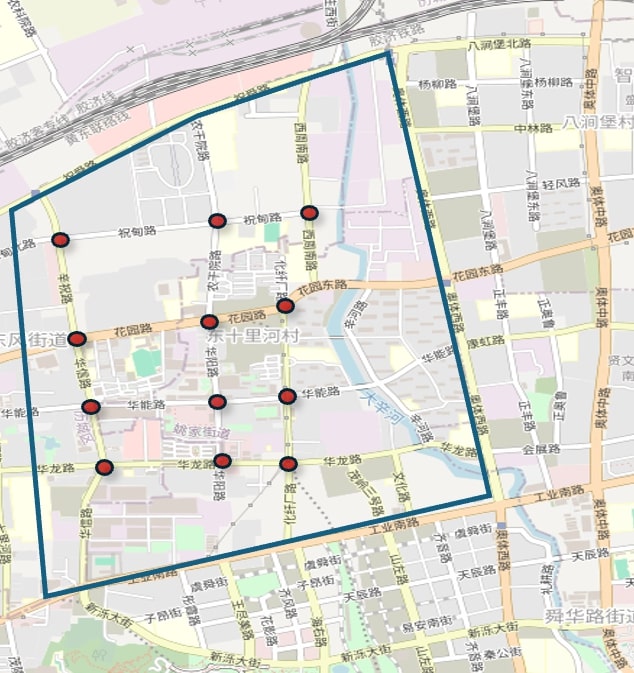}
    }
    \subfloat[Hangzhou]{
        \label{fig:hangzhou}
        \includegraphics[width=0.25\textwidth]{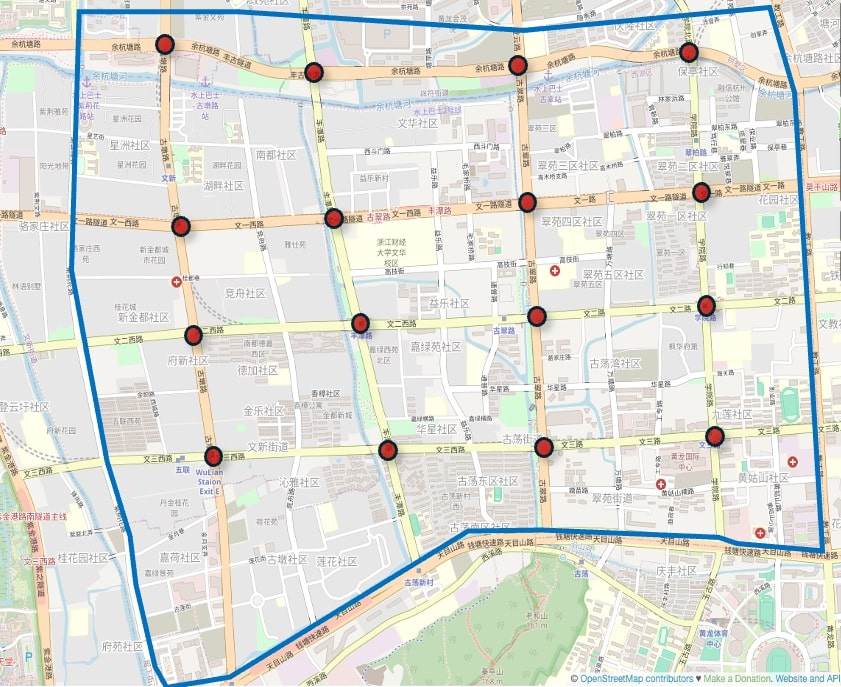}
    }
    \subfloat[Manhattan]{
        \label{fig:manhattan}
        \includegraphics[width=0.25\textwidth]{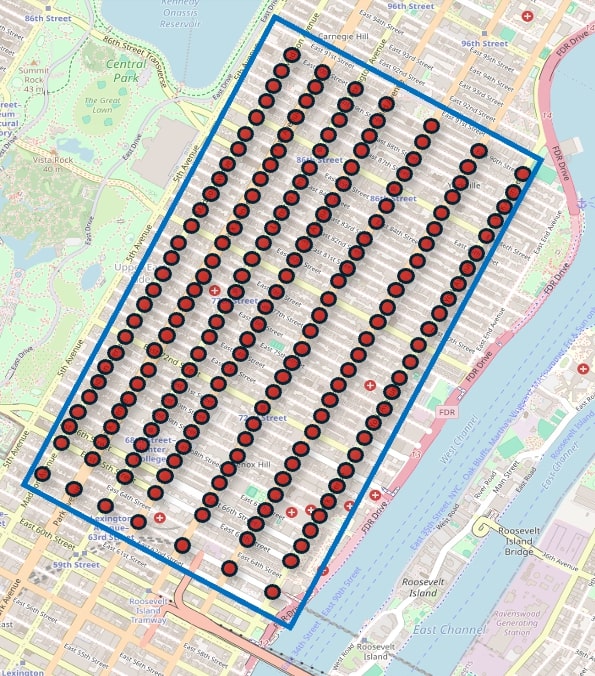}
    }
    \caption{Traffic networks used in the experiments: (a) Jinan, (b) Hangzhou, and (c) Manhattan, illustrating the varying scales and complexities of the test scenarios.}
    \label{fig:scenarios}
    \vspace{-1.5em}
\end{figure*}

OffLight integrates IS and RBPS to improve sample efficiency and mitigate distributional shift in offline MARL for TSC. IS ensures transitions align with the target policy, while RBPS prioritizes high-reward episodes, refining learning by emphasizing both relevance and quality\footnote{Multiplying \( w_{\text{IS}, t}^{k} \) and \( w_{\text{RBPS}}^{k} \) ensures that only transitions that are both relevant to the target policy and come from successful episodes are given significant weight. This dual prioritization enhances the effectiveness and robustness of policy learning.}.
\begin{equation}
    w_{\text{combined}, t}^{k} = w_{\text{IS}, t}^{k} \times w_{\text{RBPS}}^{k}
\end{equation}

To maintain stability, the combined weights are normalized and clipped within each minibatch. These adjusted weights are then incorporated into the Temporal Difference (TD) error and Conservative Q-Learning (CQL) loss functions, ensuring more effective policy learning. 

\subsection{Integration with Offline RL Algorithms}
\label{sec:integration_offline_rl}

OffLight can easily be integrated with existing offline RL algorithms. CQL and TD3+BC are chosen as baseline offline RL algorithms. The combined weighting scheme—comprising IS weights and RBPS weights—is incorporated into both the loss functions and the sampling mechanisms of these algorithms to ensure effective bias correction and sample prioritization.

\noindent \textbf{Loss Function Adjustment}: The combined weights \( \tilde{w}_{\text{combined}, t} \) are used to scale the loss components, ensuring that transitions from high-reward episodes aligned with the target policy have a greater impact on policy and value updates. This scaling is formalized in the loss functions of CQL and TD3+BC as follows:
\begin{equation}
    \mathcal{L} = \mathbb{E}_{(\mathcal{O}, \mathcal{A}, \mathcal{O}', \mathcal{R}) \sim \mathcal{D}} \left[ \tilde{w}_{\text{combined}, t} \cdot \ell(\mathcal{O}, \mathcal{A}, \mathcal{O}', \mathcal{R}) \right],
\end{equation}
where \( \ell(\mathcal{O}, \mathcal{A}, \mathcal{O}', \mathcal{R}) \) represents the specific loss term associated with the chosen RL algorithm (e.g., Bellman error for CQL or policy loss for TD3+BC).

\noindent \textbf{Sampling Mechanism}: Episodes are sampled based on their total rewards, prioritizing those with higher cumulative returns. This is implemented by adjusting the sampling probabilities \( p_i \) according to \( w_{\text{RBPS}} \), ensuring that more successful episodes are sampled more frequently. This prioritization enhances the efficiency of the learning process by focusing on the most informative and effective experiences.

\section{Experimental Setup}
\label{sec:experimental_setup}

We evaluate OffLight in three real-world urban traffic scenarios \footnote{\url{https://github.com/traffic-signal-control/sample-code/}}: Jinan (12 intersections), Hangzhou (16 intersections), and Manhattan (196 intersections). These scenarios vary in size and traffic demand, allowing for a comprehensive assessment of OffLight’s scalability and robustness. Traffic demand is classified as low, medium (real-world data), or high by adjusting the number of vehicles entering the network proportionally.

The experiments are designed to address the following key questions:
\begin{enumerate}
    \item How effectively does OffLight model heterogeneous behavior policies in offline datasets?
    \item What is the impact of incorporating importance sampling on learning from mixed-policy datasets?
    \item How does the proportion of suboptimal policy data affect overall performance?
\end{enumerate}

\begin{table}[!h]    
    \resizebox{\columnwidth}{!}{%
        \begin{tabular}{@{}ccccc@{}}            
            \toprule
            \begin{tabular}[c]{@{}l@{}}
                Average Network \\ Demand (veh/hr)
            \end{tabular} 
            & Low & Medium & High & Intersections \\ \midrule
            Jinan & 450 & 550 & 650 & 12 \\
            Hangzhou & 350 & 420 & 500 & 16 \\
            Manhattan & 150 & 250 & 350 & 196 \\ \bottomrule
        \end{tabular}%
    } 
    \caption{Traffic demand levels and network sizes for Jinan, Hangzhou, and Manhattan. Demand represents vehicles generated per incoming lane per hour.}
    \label{tab:scenarios}
    \vspace{-1.5em}
\end{table}

\noindent \textbf{Traffic Scenarios:} The selected networks represent different levels of complexity: Jinan (moderate, 12 intersections), Hangzhou (higher traffic, 16 intersections), and Manhattan (large-scale, 196 intersections). Each scenario simulates low, medium, and high demand conditions (Table~\ref{tab:scenarios}). 
\linebreak
\noindent \textbf{Data Collection: } Training data includes traffic simulations using multiple controllers: rule-based methods (Fixed Time, Greedy, Max Pressure, SOTL), expert RL controllers (MAPPO), and random policies for suboptimal behavior. Each simulation runs for 1 hour, divided into 10 episodes of 6 minutes each, across three demand levels, totaling 100 simulation hours per controller per demand level.
\linebreak
\noindent \textbf{Baselines:} OffLight is compared against Behavior Cloning (BC), Conservative Q-Learning (CQL), and TD3+BC. Each algorithm is trained for 20k timesteps.
\linebreak
\noindent \textbf{Evaluation Metrics: } Performance is measured using Average Travel Time (ATT) and Queue Length (QL), assessing traffic flow and congestion reduction across different demand levels.

\section{Results and Discussion}
\label{sec:results}

This section presents the evaluation of OffLight across various traffic scenarios. We measure performance in terms of Average Travel Time (ATT) and Queue Length (QL), highlighting the effectiveness of OffLight in learning from mixed-policy datasets.

\subsection{Performance on Mixed-Quality Data}

We assess OffLight’s performance on datasets with varying proportions of expert and random policy data. The results are shown for three traffic networks: Jinan, Hangzhou, and Manhattan, under low, medium, and high traffic demand. Results show that OffLight consistently outperforms baselines, particularly in high-demand conditions, demonstrating its ability to manage congestion in large-scale networks.

\begin{table*}[!ht]
    \centering
    \resizebox{\textwidth}{!}{%
    \begin{tabular}{llccccccccc}
        \toprule
        \multirow{3}{*}{\textbf{Category}} & \multirow{3}{*}{\textbf{Algorithm}} & \multicolumn{3}{c}{\textbf{Jinan}} & \multicolumn{3}{c}{\textbf{Hangzhou}} & \multicolumn{3}{c}{\textbf{Manhattan}} \\
        \cmidrule(lr){3-5} \cmidrule(lr){6-8} \cmidrule(lr){9-11}
        &  & \textbf{Low} & \textbf{Medium} & \textbf{High} & \textbf{Low} & \textbf{Medium} & \textbf{High} & \textbf{Low} & \textbf{Medium} & \textbf{High} \\
        \midrule
        \multirow{5}{*}{\textit{Rule-based}} & Random & 411.97 & 450.61 & 470.82 & 418.89 & 430.01 & 451.15 & 774.33 & 784.56 & 790.14 \\
        & Fixed Time & 354.95 & 380.26 & 407.25 & 370.11 & 392.69 & 407.33 & 539.07 & 574.17 & 612.77 \\
        & Greedy & 287.10 & 310.08 & 337.60 & 303.02 & 307.41 & 314.72 & 484.01 & 527.01 & 575.03 \\
        & Max Pressure & 386.43 & 409.47 & 434.91 & 405.95 & 427.04 & 437.44 & 640.96 & 658.31 & 684.59 \\
        & SOTL & 399.30 & 418.34 & 443.35 & 447.22 & 470.14 & 476.91 & 540.55 & 583.13 & 615.19 \\
        \midrule
        \multirow{1}{*}{\textit{Online}} & MAPPO & 263.87 & 282.86 & 301.49 & 272.57 & 280.82 & 292.72 & 322.99 & 367.83 & 402.99 \\
        \midrule
        \multirow{5}{*}{\textit{Offline}} & BC & 364.50 & 388.18 & 409.36 & 382.59 & 393.38 & 408.46 & 561.07 & 595.22 & 617.59 \\
        & TD3+BC & 321.87 & 356.51 & 364.26 & 324.18 & 338.52 & 364.97 & 489.78 & 495.13 & 545.11 \\
        & CQL & 326.06 & 337.74 & 372.41 & 369.63 & 350.06 & 367.33 & 495.29 & 524.25 & 582.87 \\
        & OffLight (TD3+BC) & \cellcolor{lightblue}292.90 & 317.29 & \cellcolor{lightblue}316.91 & \cellcolor{lightblue}288.52 & 299.59 & \cellcolor{lightblue}313.87 & \cellcolor{lightblue}440.80 & 423.34 & \cellcolor{lightblue}441.54 \\
        & OffLight (CQL) & 303.24 & \cellcolor{lightblue}307.34 & 324.00 & 325.27 & \cellcolor{lightblue}304.55 & 308.56 & 439.82 & \cellcolor{lightblue}456.10 & 483.78 \\
        \bottomrule
    \end{tabular}
    }
    \caption{Average Travel Time (ATT): Mixed Policy Dataset}
    \label{tab:att_mixed}
\end{table*}

\begin{table*}[!ht]
    \centering
    \resizebox{\textwidth}{!}{%
    \begin{tabular}{llccccccccc}
        \toprule
        \multirow{3}{*}{\textbf{Category}} & \multirow{3}{*}{\textbf{Algorithm}} & \multicolumn{3}{c}{\textbf{Jinan}} & \multicolumn{3}{c}{\textbf{Hangzhou}} & \multicolumn{3}{c}{\textbf{Manhattan}} \\
        \cmidrule(lr){3-5} \cmidrule(lr){6-8} \cmidrule(lr){9-11}
        &  & \textbf{Low} & \textbf{Medium} & \textbf{High} & \textbf{Low} & \textbf{Medium} & \textbf{High} & \textbf{Low} & \textbf{Medium} & \textbf{High} \\
        \midrule
        \multirow{5}{*}{\textit{Rule-based}} & Random & 590.14 & 809.45 & 944.45 & 260.09 & 295.46 & 357.30 & 2899.23 & 3291.49 & 3590.03 \\
        & Fixed Time & 411.88 & 553.15 & 716.47 & 193.45 & 231.09 & 270.04 & 615.94 & 757.01 & 884.03 \\
        & Greedy & 216.56 & 356.59 & 514.37 & 170.81 & 192.55 & 219.85 & 512.27 & 676.02 & 830.96 \\
        & Max Pressure & 500.51 & 639.20 & 803.91 & 254.08 & 309.32 & 343.49 & 758.94 & 881.23 & 991.15 \\
        & SOTL & 518.57 & 645.27 & 790.08 & 320.91 & 377.54 & 418.35 & 607.80 & 767.82 & 886.84 \\
        \midrule
        \multirow{1}{*}{\textit{Online}} & MAPPO & 170.98 & 213.19 & 228.95 & 152.81 & 162.34 & 171.91 & 374.44 & 452.74 & 512.33 \\
        \midrule
        \multirow{5}{*}{\textit{Offline}} & BC & 417.35 & 554.59 & 689.67 & 233.26 & 269.77 & 308.28 & 971.47 & 1173.25 & 1303.26 \\
        & TD3+BC & 372.31 & 509.34 & 586.35 & 193.78 & 240.24 & 273.07 & 874.53 & 956.15 & 1129.03 \\
        & CQL & 369.72 & 477.32 & 566.41 & 227.52 & 240.24 & 275.83 & 864.52 & 1023.94 & 1218.43 \\
        & OffLight (TD3+BC) & \cellcolor{lightblue}312.88 & 423.88 & \cellcolor{lightblue}493.12 & \cellcolor{lightblue}169.07 & 222.17 & \cellcolor{lightblue}222.61 & 806.83 & \cellcolor{lightblue}944.64 & \cellcolor{lightblue}962.81 \\
        & OffLight (CQL) & 314.86 & \cellcolor{lightblue}402.17 & 553.09 & 189.50 & \cellcolor{lightblue}219.46 & 240.54 & \cellcolor{lightblue}798.09 & 979.13 & 999.89 \\
        \bottomrule
        \end{tabular}
    }
    \caption{Queue Length (QL): Mixed Policy Dataset}
    \label{tab:queue_length_mixed}
\end{table*}

\begin{figure}[!h]
    \centering
    \includegraphics[width=\linewidth]{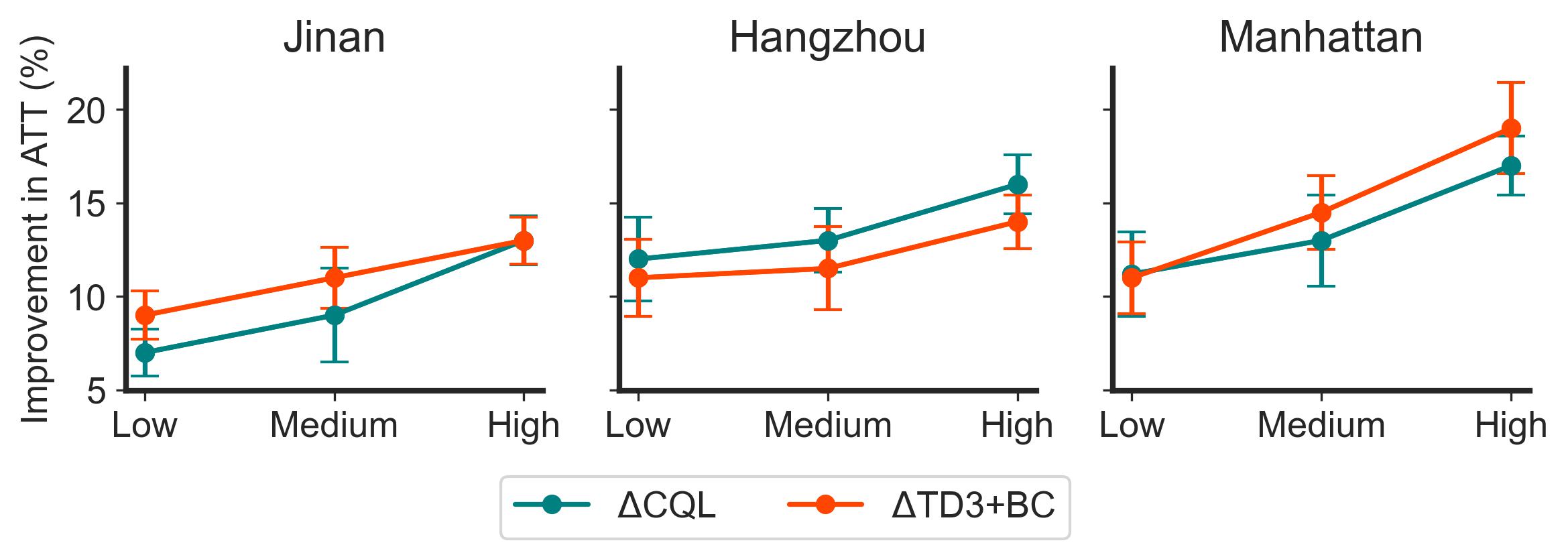}
    \caption{Performance Comparison of OffLight (CQL) and OffLight (TD3+BC) on Mixed Data showing the improvements in average travel time (ATT) across different traffic demand levels}
    \label{fig:mixed_att_comparison}
    \vspace{-1.5em}
\end{figure}

\subsubsection{Average Travel Time (ATT)} 

Figure \ref{fig:mixed_att_comparison} compares OffLight to offline RL baselines. OffLight achieves lower ATT across all scenarios, with the greatest improvements in high-demand conditions. In Manhattan, OffLight reduces ATT by up to 7.8\% (TD3+BC variant) and 6.9\% (CQL variant) compared to their respective baselines, proving its scalability in complex networks. Performance gains are smaller in low-traffic scenarios, where traffic flow is already manageable. These results indicate that OffLight effectively prioritizes high-quality data, improving traffic efficiency in congested environments.

\subsubsection{Queue Length (QL)} 

\begin{figure}[!h]
    \centering
    \includegraphics[width=0.7\linewidth]{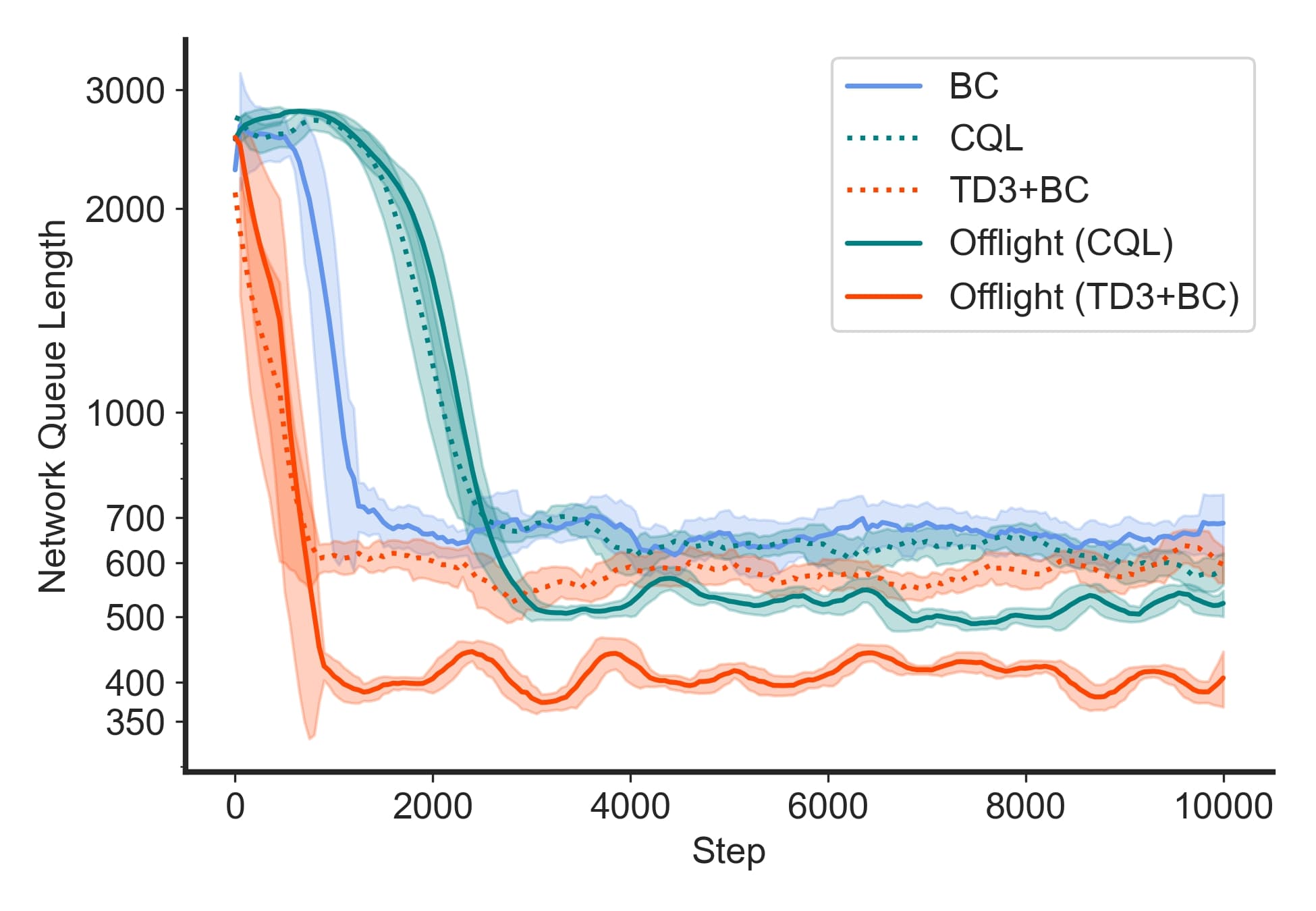}
    \caption{Learning curves for the Jinan scenario under medium traffic demand, showing the convergence of queue length for OffLight and baseline methods.}
    \label{fig:learning_curves}
    \vspace{-1em}
\end{figure}

Table \ref{tab:queue_length_mixed} and Figure \ref{fig:learning_curves} show that OffLight significantly reduces queue lengths, especially in medium and high traffic conditions. In Hangzhou, OffLight (TD3+BC) decreases queue lengths by 11.2\% compared to TD3+BC alone. In Manhattan, OffLight (CQL) achieves a 6\% reduction under high traffic demand, demonstrating its capability to optimize traffic flow at intersections. Improvements are smaller in low-traffic scenarios where congestion is naturally less severe.

Figure \ref{fig:learning_curves} highlights the learning curves for the Jinan scenario under medium traffic demand. OffLight (TD3+BC) and OffLight (CQL) achieve faster convergence and lower queue lengths compared to baselines, with OffLight (TD3+BC) performing best. TD3+BC and CQL improve over time but maintain higher queue lengths than OffLight, while BC and other baselines show greater variance and slower convergence.

OffLight’s IS mechanism ensures that transitions are weighted based on their relevance to the learned policy, reducing reliance on low-quality data. This re-weighting helps OffLight make better traffic control decisions, resulting in smoother traffic flow, quicker convergence, and significantly lower queue lengths across all demand levels.
    
\subsubsection{Behavior Representation} 

\begin{figure}[!h]
    \centering
    \includegraphics[width=\linewidth]{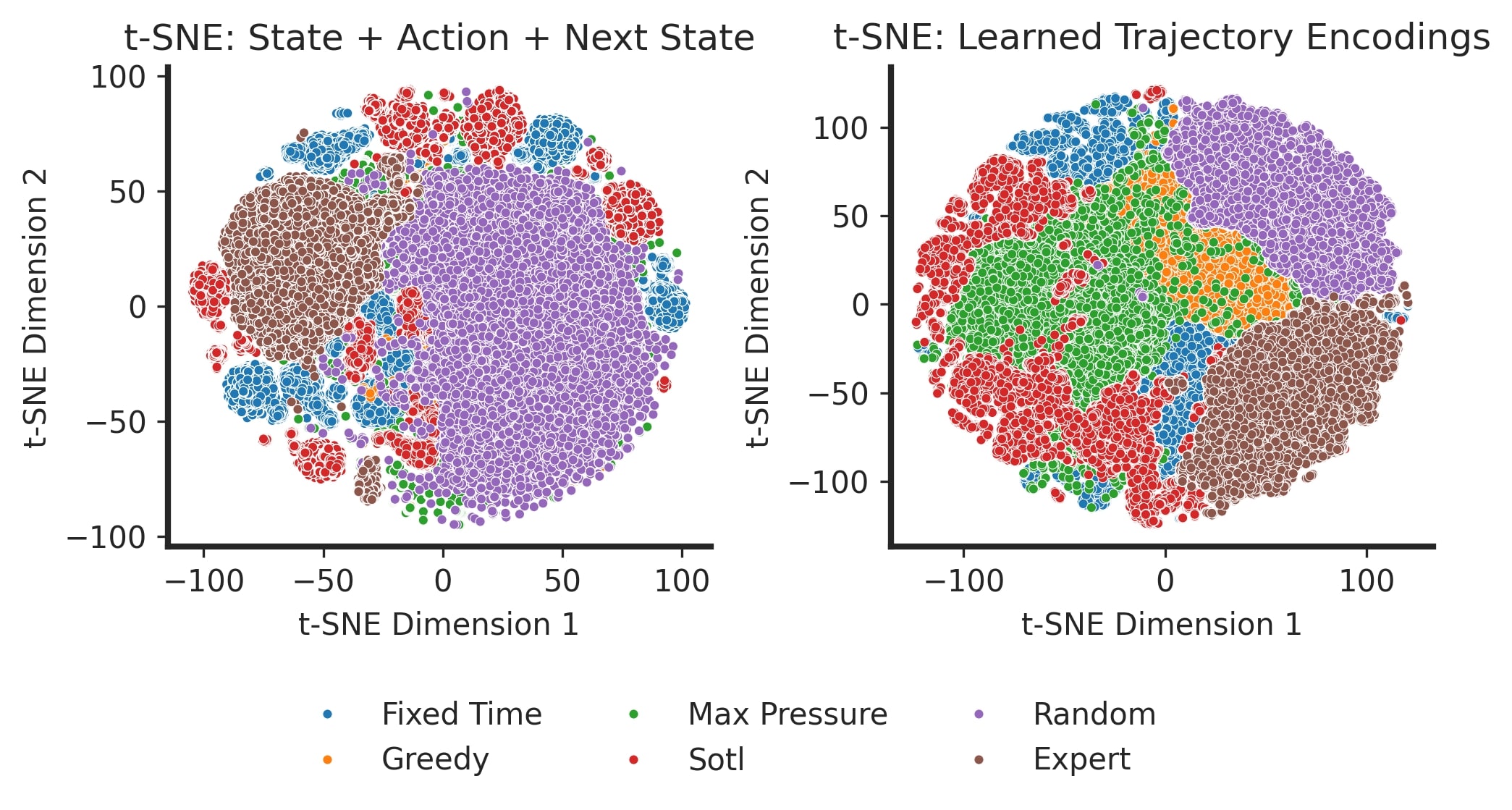}
    \caption{t-SNE Representation of Actual Data and Learned Trajectories for Hangzhou scenario}
    \label{fig:mixed_clustering}
    \vspace{-1.5em}
\end{figure}

Figure \ref{fig:mixed_clustering} visualizes t-SNE embeddings of learned latent space in the Hangzhou scenario, illustrating OffLight’s ability to differentiate between expert, random, and rule-based policies. The distinct clusters confirm that GMM-VGAE effectively captures structural differences in control policies, enabling more robust policy learning.

OffLight remains effective even as the proportion of expert data decreases, consistently outperforming baselines. This robustness is crucial in real-world settings where dataset quality varies across intersections and traffic conditions, ensuring reliable performance despite heterogeneous data sources.

\subsection{Ablation Studies}

\subsubsection{Performance Comparison with Different Levels of Mixing}

\begin{figure}[!h]
    \centering
    \includegraphics[width=\linewidth]{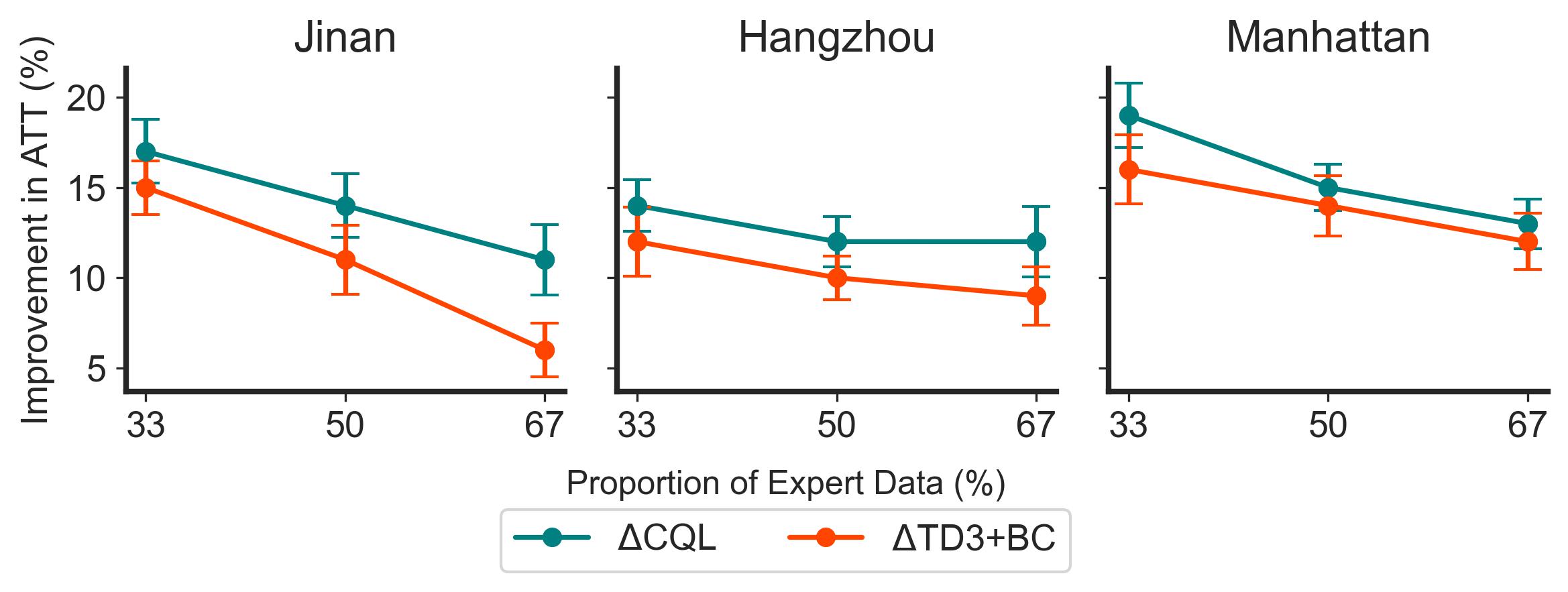}
    \caption{Performance Comparison of OffLight with Expert and Random Data Mix}
    \label{fig:comparison_plots_1}
    \vspace{-1.5em}
\end{figure}

In this study, we analyze the impact of mixing varying proportions of data from expert and random policies on the performance of the OffLight framework compared to conventional offline RL algorithms. Specifically, we explore how OffLight performs under different dataset compositions: 33\%, 50\%, and 67\% expert policy data, with the remainder coming from random policies.
    
Figure~\ref{fig:comparison_plots_1} presents the performance comparison across different datasets. As the proportion of expert data decreases and random policy data increases, the performance of all algorithms degrades, which is expected. However, OffLight exhibits a more gradual performance decline compared to TD3+BC and CQL. This indicates that OffLight’s IS and RBPS mechanisms are effective in mitigating the negative effects of suboptimal data.

\subsubsection{Effectiveness of Improved Sampling Strategies}

\begin{figure}[!h]
    \centering
    \includegraphics[width=\linewidth]{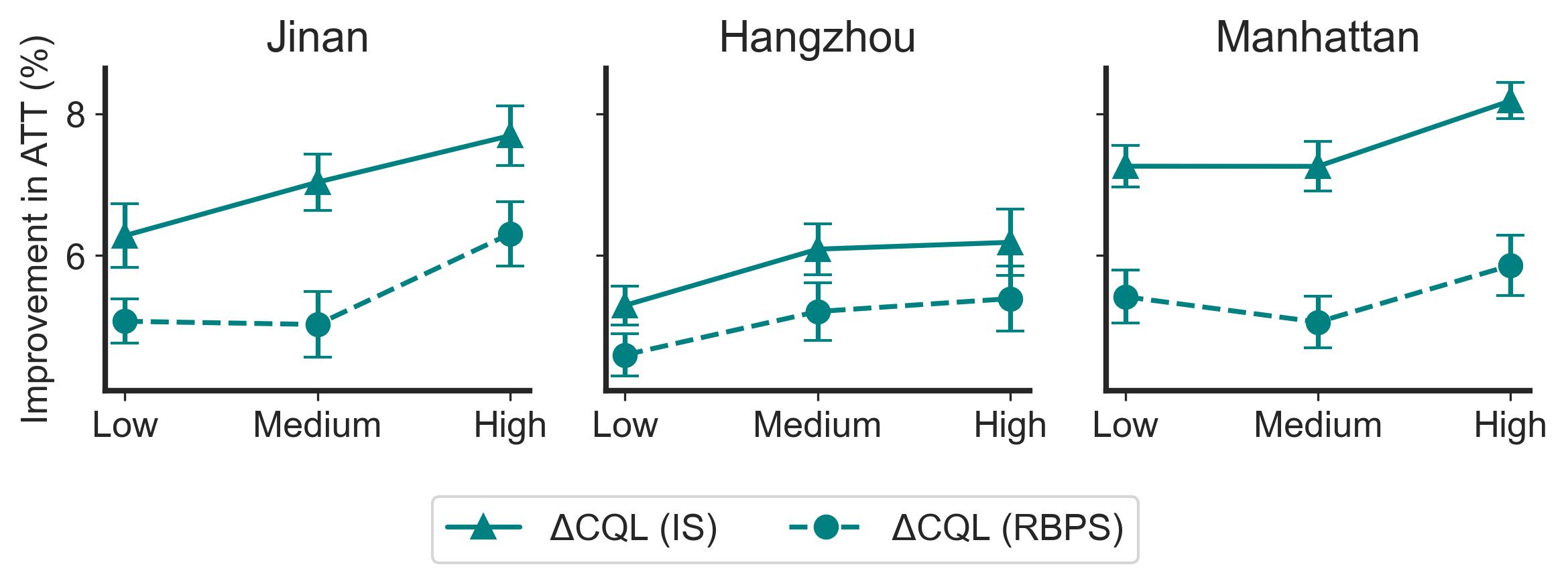}
    \vspace{1em} 
    \includegraphics[width=\linewidth]{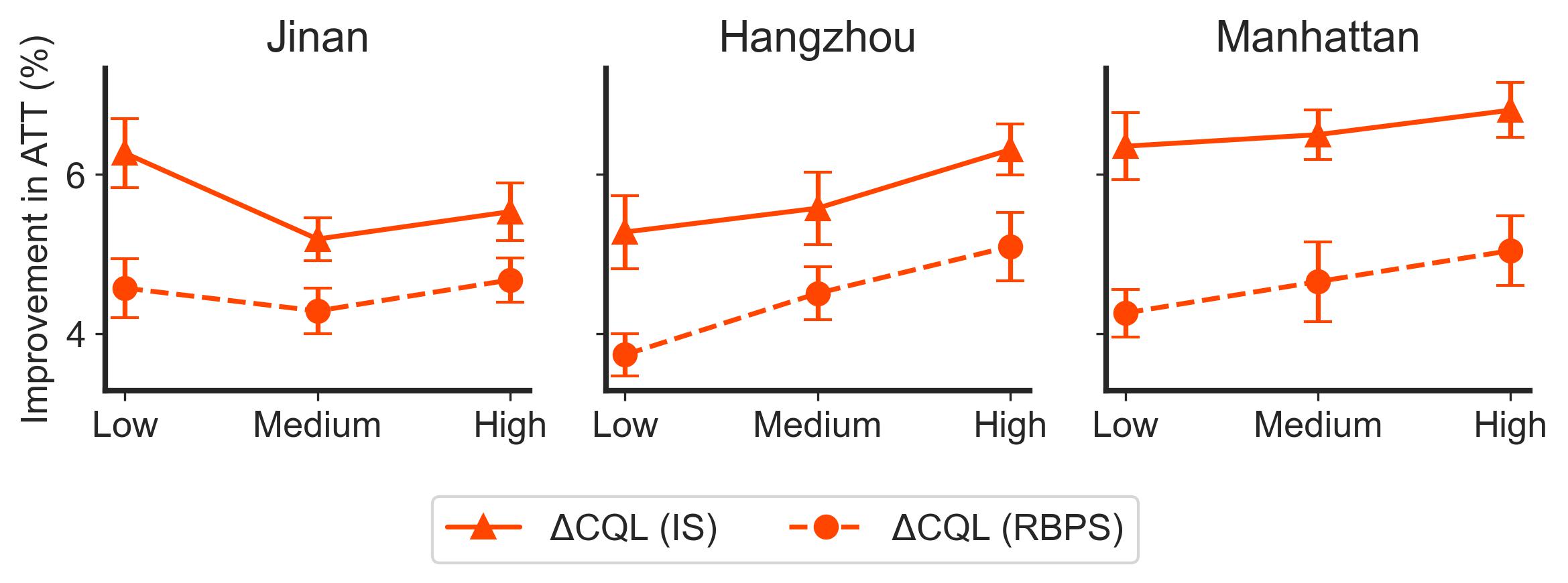}
    \caption{Ablation Study on CQL (top) and TD3+BC (bottom) with IS and RBPS}
    \label{fig:comparison_plots_2}
    \vspace{-1.em}
\end{figure}

Figure \ref{fig:comparison_plots_2} analyzes the impact of IS and RBPS on OffLight’s performance. IS improves performance more than RBPS across all scenarios and traffic demand levels, with the performance gap widening as congestion increases. In Manhattan’s high-demand setting, IS improves ATT by up to 10\% for CQL and 9\% for TD3+BC, demonstrating its effectiveness in mitigating distributional shift. RBPS provides smaller, incremental gains (4-6\% ATT reduction), especially in medium and high-demand settings. While RBPS prioritizes high-reward episodes, it is more sensitive to noisy data and less effective than IS in complex networks like Manhattan.

By re-weighting transitions based on their relevance to the target policy, IS ensures that OffLight learns from high-quality data, improving traffic flow decisions in congested scenarios. RBPS, while useful, has a smaller impact when suboptimal policies dominate certain episodes. The combined approach enhances OffLight’s ability to learn from offline datasets efficiently, particularly in large-scale networks.

\subsection{Discussion}

This study introduced OffLight, an offline MARL framework for traffic signal control (TSC) using heterogeneous datasets. OffLight consistently outperforms baselines, demonstrating robustness across different traffic conditions and dataset compositions. By leveraging GMM-VGAE for behavior modeling and incorporating IS and RBPS, OffLight effectively handles mixed-quality data, prioritizing high-reward transitions and mitigating distributional shifts.

OffLight scales well across diverse network sizes, from small to large-scale urban environments. However, its reliance on GMM-VGAE introduces computational overhead, which may limit real-time deployment. Additionally, its performance depends on dataset quality, requiring a balanced mix of expert and suboptimal policies for optimal learning.

Beyond TSC, OffLight’s framework is applicable to smart grids \cite{vamvakas2023review} and supply chain management \cite{mousa2024analysis}, where offline MARL can optimize decision-making in networked systems. Its IS-based policy correction and graph-based modeling make it well-suited for improving coordination in distributed environments.

\section{Conclusion}
\label{sec:conclusion}

OffLight demonstrates the potential to significantly enhance traffic signal control (TSC) in urban networks through offline multi-agent reinforcement learning (MARL). By addressing the key challenges of heterogeneous behavior policies and distributional shifts, OffLight offers a robust solution capable of learning effective policies from mixed-quality datasets. The framework's incorporation of GMM-VGAE for behavior policy modeling, alongside IS and RBPS, leads to substantial improvements in traffic efficiency. These advancements reflect the ability of OffLight to scale effectively across different network sizes and traffic conditions, making it a practical tool for offline reinforcement learning for real-world traffic management. 

\section*{APPENDIX}

\subsection{Importance Sampling Weight Consistency}
\label{sec:appendix_is_consistency}

In OffLight, IS is integrated to correct for the distributional shift between the behavior policy \( \pi_b \) and the target policy \( \pi_{\theta} \). We aim to show that the importance sampling weights computed using the estimated behavior policy \( \hat{\pi}_b \) provide consistent and unbiased estimates of expectations under the target policy. For any function \( f(\tau) \) of a trajectory \( \tau \), the expectation under \( \pi_{\theta} \) can be estimated using samples from \( \pi_b \) as:
\begin{equation}
    \mathbb{E}_{\tau \sim \pi_{\theta}} [f(\tau)] = \mathbb{E}_{\tau \sim \pi_b} \left[ w(\tau) f(\tau) \right],
\end{equation}
where the importance weight \( w(\tau) \) is given by:
\begin{equation}
    w(\tau) = \prod_{t=0}^{T} \prod_{n=1}^{N} \frac{\pi_{\theta}^{i}(a_t^{i} \mid s_t^{i})}{\hat{\pi}_b^{i}(a_t^{i} \mid s_t^{i})}.
\end{equation}
\textbf{Assumptions:}
\begin{enumerate}
    \item 1. The estimated behavior policy \( \hat{\pi}_b \) satisfies \( \hat{\pi}_b^{i}(a_t^{i} \mid s_t^{i}) = \pi_b^{i}(a_t^{i} \mid s_t^{i}) \) for all \( n, t \).
    \item The support of \( \pi_{\theta}^{i}(a \mid s) \) is contained within the support of \( \pi_b^{i}(a \mid s) \) for all \( n \), i.e., \( \pi_b^{i}(a \mid s) = 0 \implies \pi_{\theta}^{i}(a \mid s) = 0 \).
\end{enumerate}
\textbf{Proof:} Under these assumptions, the importance weights provide an unbiased estimate:
\begin{align}
    \mathbb{E}_{\tau \sim \pi_b} \left[ w(\tau) f(\tau) \right] &= \int_{\tau} \pi_b(\tau) w(\tau) f(\tau) \, d\tau \\
    &= \int_{\tau} \pi_b(\tau) \frac{\pi_{\theta}(\tau)}{\pi_b(\tau)} f(\tau) \, d\tau \\
    &= \int_{\tau} \pi_{\theta}(\tau) f(\tau) \, d\tau \\
    &= \mathbb{E}_{\tau \sim \pi_{\theta}} [f(\tau)].
\end{align}
This demonstrates that OffLight's IS weights yield consistent estimates when the behavior policy is accurately estimated.

\subsection{Variance Reduction via Accurate Behavior Policy Estimation}
\label{sec:appendix_variance_reduction}

While IS provides unbiased estimates, the variance of the estimator can be high, particularly when the target and behavior policies differ significantly. By accurately estimating \( \pi_b \) using the GMM-VGAE, OffLight reduces the variance of the importance weights. The variance of the IS estimator is given by:
\begin{equation}
    \text{Var}_{\pi_b} [w(\tau) f(\tau)] = \mathbb{E}_{\pi_b} [w^2(\tau) f^2(\tau)] - \left( \mathbb{E}_{\pi_{\theta}} [f(\tau)] \right)^2.
\end{equation}
An accurate estimation of \( \pi_b \) ensures that \( w(\tau) \) does not have extremely large values, which would otherwise inflate the variance. Specifically, when \( \pi_{\theta} \) is close to \( \pi_b \), \( w(\tau) \) is close to 1, minimizing the variance.

\noindent \textbf{Theorem:} If \( \hat{\pi}_b \) is a consistent estimator of \( \pi_b \), then as the amount of data increases, the variance of the IS estimator decreases.

\noindent \textbf{Proof:} As \( \hat{\pi}_b \) converges to \( \pi_b \), the distribution of \( w(\tau) \) becomes tighter around 1, reducing the variance. This is formalized by the Law of Large Numbers and convergence properties of consistent estimators.


\bibliography{references}

\begin{thebibliography}{10}

\bibitem{haydari2020deep}
A.~Haydari and Y.~Y{\i}lmaz, ``Deep reinforcement learning for intelligent transportation systems: A survey,'' {\em IEEE Transactions on Intelligent Transportation Systems}, vol.~23, no.~1, pp.~11--32, 2020.

\bibitem{noaeen2022reinforcement}
M.~Noaeen, A.~Naik, L.~Goodman, J.~Crebo, T.~Abrar, Z.~S.~H. Abad, A.~L. Bazzan, and B.~Far, ``Reinforcement learning in urban network traffic signal control: A systematic literature review,'' {\em Expert Systems with Applications}, vol.~199, p.~116830, 2022.

\bibitem{levine2020offline}
S.~Levine, A.~Kumar, G.~Tucker, and J.~Fu, ``Offline reinforcement learning: Tutorial, review, and perspectives on open problems,'' {\em arXiv preprint arXiv:2005.01643}, 2020.

\bibitem{zhan2022offline}
W.~Zhan, B.~Huang, A.~Huang, N.~Jiang, and J.~Lee, ``Offline reinforcement learning with realizability and single-policy concentrability,'' in {\em Conference on Learning Theory}, pp.~2730--2775, PMLR, 2022.

\bibitem{fu2020d4rl}
J.~Fu, A.~Kumar, O.~Nachum, G.~Tucker, and S.~Levine, ``D4rl: Datasets for deep data-driven reinforcement learning,'' {\em arXiv preprint arXiv:2004.07219}, 2020.

\bibitem{lange2012batch}
S.~Lange, T.~Gabel, and M.~Riedmiller, ``Batch reinforcement learning,'' in {\em Reinforcement learning: State-of-the-art}, pp.~45--73, Springer, 2012.

\bibitem{xiong2019learning}
Y.~Xiong, G.~Zheng, K.~Xu, and Z.~Li, ``Learning traffic signal control from demonstrations,'' in {\em Proceedings of the 28th ACM international conference on information and knowledge management}, pp.~2289--2292, 2019.

\bibitem{huo2020cooperative}
Y.~Huo, Q.~Tao, and J.~Hu, ``Cooperative control for multi-intersection traffic signal based on deep reinforcement learning and imitation learning,'' {\em Ieee Access}, vol.~8, pp.~199573--199585, 2020.

\bibitem{sun2024crosslight}
Q.~Sun, R.~Zha, L.~Zhang, J.~Zhou, Y.~Mei, Z.~Li, and H.~Xiong, ``Crosslight: Offline-to-online reinforcement learning for cross-city traffic signal control,'' in {\em Proceedings of the 30th ACM SIGKDD Conference on Knowledge Discovery and Data Mining}, pp.~2765--2774, 2024.

\bibitem{chen2022real}
R.~Chen, F.~Fang, and N.~Sadeh, ``The real deal: A review of challenges and opportunities in moving reinforcement learning-based traffic signal control systems towards reality,'' {\em arXiv preprint arXiv:2206.11996}, 2022.

\bibitem{wei2019colight}
H.~Wei, N.~Xu, H.~Zhang, G.~Zheng, X.~Zang, C.~Chen, W.~Zhang, Y.~Zhu, K.~Xu, and Z.~Li, ``Colight: Learning network-level cooperation for traffic signal control,'' in {\em Proceedings of the 28th ACM international conference on information and knowledge management}, pp.~1913--1922, 2019.

\bibitem{kumar2020conservative}
A.~Kumar, A.~Zhou, G.~Tucker, and S.~Levine, ``Conservative q-learning for offline reinforcement learning,'' {\em Advances in Neural Information Processing Systems}, vol.~33, pp.~1179--1191, 2020.

\bibitem{wu2019behavior}
Y.~Wu, G.~Tucker, and O.~Nachum, ``Behavior regularized offline reinforcement learning,'' {\em arXiv preprint arXiv:1911.11361}, 2019.

\bibitem{hong2023beyond}
Z.-W. Hong, A.~Kumar, S.~Karnik, A.~Bhandwaldar, A.~Srivastava, J.~Pajarinen, R.~Laroche, A.~Gupta, and P.~Agrawal, ``Beyond uniform sampling: Offline reinforcement learning with imbalanced datasets,'' {\em Advances in Neural Information Processing Systems}, vol.~36, pp.~4985--5009, 2023.

\bibitem{yue2023decoupled}
Y.~Yue, B.~Kang, X.~Ma, Q.~Yang, G.~Huang, S.~Song, and S.~Yan, ``Decoupled prioritized resampling for offline rl,'' {\em arXiv preprint arXiv:2306.05412}, 2023.

\bibitem{oliehoek2016concise}
F.~A. Oliehoek, C.~Amato, {\em et~al.}, {\em A concise introduction to decentralized POMDPs}, vol.~1.
\newblock Springer, 2016.

\bibitem{mnih2015human}
V.~Mnih, K.~Kavukcuoglu, D.~Silver, A.~A. Rusu, J.~Veness, M.~G. Bellemare, A.~Graves, M.~Riedmiller, A.~K. Fidjeland, G.~Ostrovski, {\em et~al.}, ``Human-level control through deep reinforcement learning,'' {\em nature}, vol.~518, no.~7540, pp.~529--533, 2015.

\bibitem{wei2019presslight}
H.~Wei, C.~Chen, G.~Zheng, K.~Wu, V.~Gayah, K.~Xu, and Z.~Li, ``Presslight: Learning max pressure control to coordinate traffic signals in arterial network,'' in {\em Proceedings of the 25th ACM SIGKDD international conference on knowledge discovery \& data mining}, pp.~1290--1298, 2019.

\bibitem{oroojlooy2020attendlight}
A.~Oroojlooy, M.~Nazari, D.~Hajinezhad, and J.~Silva, ``Attendlight: Universal attention-based reinforcement learning model for traffic signal control,'' {\em Advances in Neural Information Processing Systems}, vol.~33, pp.~4079--4090, 2020.

\bibitem{velivckovic2017graph}
P.~Veli{\v{c}}kovi{\'c}, G.~Cucurull, A.~Casanova, A.~Romero, P.~Lio, and Y.~Bengio, ``Graph attention networks,'' {\em arXiv preprint arXiv:1710.10903}, 2017.

\bibitem{kumar2019stabilizing}
A.~Kumar, J.~Fu, M.~Soh, G.~Tucker, and S.~Levine, ``Stabilizing off-policy q-learning via bootstrapping error reduction,'' {\em Advances in neural information processing systems}, vol.~32, 2019.

\bibitem{fujimoto2021minimalist}
S.~Fujimoto and S.~S. Gu, ``A minimalist approach to offline reinforcement learning,'' {\em Advances in neural information processing systems}, vol.~34, pp.~20132--20145, 2021.

\bibitem{marl-book}
S.~V. Albrecht, F.~Christianos, and L.~Sch\"afer, {\em Multi-Agent Reinforcement Learning: Foundations and Modern Approaches}.
\newblock MIT Press, 2024.

\bibitem{foerster2018counterfactual}
J.~Foerster, G.~Farquhar, T.~Afouras, N.~Nardelli, and S.~Whiteson, ``Counterfactual mlti-agent policy gradients,'' in {\em Proceedings of the AAAI conference on artificial intelligence}, vol.~32, 2018.

\bibitem{vamvakas2023review}
D.~Vamvakas, P.~Michailidis, C.~Korkas, and E.~Kosmatopoulos, ``Review and evaluation of reinforcement learning frameworks on smart grid applications,'' {\em Energies}, vol.~16, no.~14, p.~5326, 2023.

\bibitem{mousa2024analysis}
M.~Mousa, D.~van~de Berg, N.~Kotecha, E.~A. del Rio~Chanona, and M.~Mowbray, ``An analysis of multi-agent reinforcement learning for decentralized inventory control systems,'' {\em Computers \& Chemical Engineering}, vol.~188, p.~108783, 2024.

\end{thebibliography}
\bibliographystyle{ieeetr}

\end{document}